\pdfoutput=1
\documentclass[11pt]{article}

\usepackage[]{acl}

\usepackage{times}
\usepackage{latexsym}

\usepackage[T1]{fontenc}
\usepackage{amsfonts}
\usepackage[utf8]{inputenc}

\usepackage{microtype}

\usepackage{amsmath}
\usepackage{amssymb}
\usepackage{graphicx}

\usepackage{float}
\usepackage{pgfplots}
\usepackage{subfigure}
\usepackage{booktabs}
\usepackage{multicol}
\usepackage{multirow}
\usepackage{arydshln} 
\usepackage{CJKutf8}
\title{Evaluate Confidence Instead of Perplexity for Zero-shot \\ Commonsense Reasoning}

\date{}

\author{Letian Peng$^{1,2,\dag}$, Zuchao Li$^{1,2,\dag}$, and Hai Zhao$^{1,2}$\thanks{$\ $  Corresponding author. $^\dag$ These authors made equal contribution. This work was supported by Key Projects of National Natural Science Foundation of China (U1836222 and
61733011).}\\
$^{1}$Department of Computer Science and Engineering, Shanghai Jiao Tong University \\
	$^{2}$MoE Key Lab of Artificial Intelligence, AI Institute, Shanghai Jiao Tong University \\
  {\tt \small \{zxc-00,charlee\}@sjtu.edu.cn, zhaohai@cs.sjtu.edu.cn}}

\begin{document}
\maketitle
\begin{abstract}

Commonsense reasoning is an appealing topic in natural language processing (NLP) as it plays a fundamental role in supporting the human-like actions of NLP systems. 
With large-scale language models as the backbone, unsupervised pre-training on numerous corpora shows the potential to capture commonsense knowledge. 
Current pre-trained language model (PLM)-based reasoning follows the traditional practice using perplexity metric. 
However, commonsense reasoning is more than existing probability evaluation, which is biased by word frequency. 
This paper reconsiders the nature of commonsense reasoning and proposes a novel commonsense reasoning metric, Non-Replacement Confidence (NRC). 
In detail, it works on PLMs according to the Replaced Token Detection (RTD) pre-training objective in ELECTRA, in which the corruption detection objective reflects the confidence on contextual integrity that is more relevant to commonsense reasoning than existing probability. Our proposed novel method boosts zero-shot performance on two commonsense reasoning benchmark datasets and further seven commonsense question-answering datasets. 
Our analysis shows that pre-endowed commonsense knowledge, especially for RTD-based PLMs, is essential in downstream reasoning.

\end{abstract}

\section{Introduction}

Commonsense reasoning is the underlying basis for human-like natural language understanding of machines. Commonsense knowledge endows natural language processing (NLP) systems with the awareness of implicit background for how human inference deals with the physical world. External commonsense knowledge created by human has been successfully applied to refine NLP systems like dialogue \cite{DBLP:conf/emnlp/ZhouJCLPR21} and generation \cite{DBLP:conf/emnlp/ChakrabartyTM21}. 

\begin{figure}
    \centering
    \includegraphics[width=0.40\textwidth]{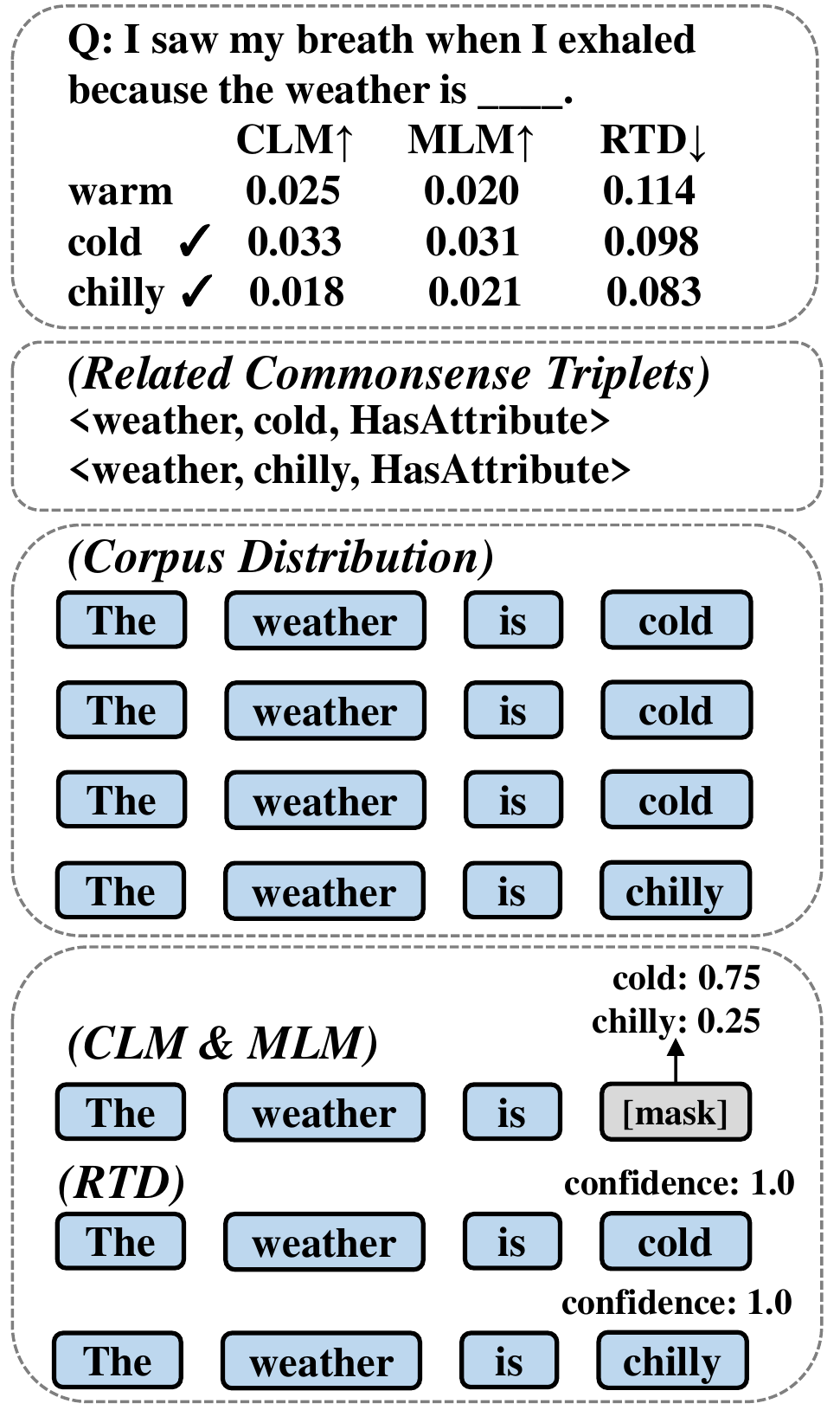}
    \caption{An instance borrowed from \cite{DBLP:conf/acl/NiuHLC0H20} that shows the bias of PLM-based inference to high-frequency words.}
    \label{fig:inst}
\end{figure}

As handcrafted commonsense dataset requires much time and energy of human annotators, many researches turn to retrieve commonsense knowledge from existing language systems. Large-scale pre-trained language models (PLMs) are desirable for the retrieval as they have been pre-trained on a wide variety of corpora to learn the interdependency between tokens. 
\citet{DBLP:conf/emnlp/PetroniRRLBWM19} exploit masked language modeling (MLM) strategy on BERT \cite{DBLP:conf/naacl/DevlinCLT19} as a knowledge base. A series of works \citep{DBLP:conf/emnlp/JiangAADN20,DBLP:conf/acl/AlghanmiAS21,DBLP:conf/eacl/HeinzerlingI21} follow this process to prompt commonsense information from PLMs, including GPT-2 \cite{DBLP:conf/nips/BrownMRSKDNSSAA20} based on casual language modeling (CLM) strategy. 

While MLM and CLM are the mainstream strategies for PLM-based commonsense reasoning, there still exists a doubt whether these learning objectives are competent to fully understand commonsense knowledge during pre-training. 
\citet{DBLP:conf/acl/NiuHLC0H20} pointed out that the inference based on word retrieval PLMs (CLM, MLM) is likely to be biased by word frequency as presented in Figure~\ref{fig:inst}. 
The word frequency perturbs the inference by assigning more positive scores to high-frequency words. The perturbance even leads to a wrong inference that \textit{warm} is assigned a higher score than \textit{chilly} in the CLM scenario. 

From the view of human-beings, commonsense knowledge represents facts in the physical world, whose confidence is independent of the statistical property in the corpus. Perplexity metric, biased to word frequency in the training corpus, is inconsistent with this nature. Essentially, the problem is caused by that MLM and CLM constrains all sentence candidates to share a total probability of $1.0$. Consequently, more frequent words will take a higher proportion of the possibility. The mutually exclusive property of perplexity underestimates confidence in other candidates when high-frequency candidates exist. On the other hand, when mentioning commonsense reasoning, we refer to the confidence in the piece of knowledge rather than the existing probability of a specific textual content. We thus conclude commonsense reasoning to be a discrimination rather than a generation (CLM-based generation or MLM-based prompting), which is currently done when calculating the sentence perplexity for the inference.

Based on the conclusion, we pursue a pre-trained discriminator towards better commonsense reasoning. ELECTRA \cite{DBLP:conf/iclr/ClarkLLM20} is a PLM trained by replaced token detection (RTD) in a GAN-like scenario. The ELECTRA discriminator is trained to detect replaced tokens from an adversarial generator. While ELECTRA does not always perform better in supervised fine-tuning \cite{DBLP:conf/iclr/ClarkLLM20}, we find that the nature of discriminator enables it to achieve significantly superior performance over other PLMs on zero-shot commonsense reasoning. For inference, we propose a new metric, Non-Replacement Confidence (NRC), to evaluate the integrity of fact descriptions. 

We experiment NRC on a wide variety of commonsense-related datasets. First, we evaluate the commonsense awareness of NRC on tuple and sentence-level descriptions. Then, we apply NRC to seven downstream commonsense question answering datasets. Experiment results verify NRC to outperform perplexity-based inference by a significant gap, showing the superiority of RTD-based discriminator to capture commonsense knowledge. NRC is also efficient to calculate as it does not require mask tokens for inference. 

Our analysis further discloses whether and how commonsense understanding benefits the downstream inference. We gather evidence, including statistics and cases, to explain the underlying principle of the application of learned commonsense knowledge to infer. RTD-based inference is verified to be more critical to components interdependent by commonsense relationships, representing a more human-like reasoning procedure. 

Our contributions are summarized as follows:

\begin{itemize}
    \item We address the inconsistency of perplexity-based evaluation with commonsense reasoning and propose the RTD-based inference to instead evaluate the confidence. 
    \item We implement a new RTD-based metric, NRC, which better discriminates the commonsense integrity of fact descriptions. Experiments on commonsense reasoning and question answering verify the superiority of NRC over conventional perplexity-based inference.
    \item Further analyses show NRC to be more capable in not only commonsense reasoning but the application of knowledge for downstream inference as well. 
    
\end{itemize}

\section{Related Work}

\subsection{Commonsense Knowledge}

Commonsense knowledge, also known as background knowledge, is the underlying basis of logic in the inference of human. As commonsense knowledge is rarely expressed in textual contents \cite{DBLP:conf/cikm/GordonD13}, many datasets \citep{DBLP:conf/sigmod/BollackerEPST08,DBLP:conf/icml/NickelTK11,DBLP:journals/corr/YangYHGD14a,DBLP:conf/acl/LiTTG16} have been handcrafted to train NLP systems and endow them with the ability to make physical world-based inference.

Following the storage system in databases, commonsense knowledge is generally formalized as a tuple $(LT, RT, REL)$, e.g. ConceptNet \cite{DBLP:conf/acl/LiTTG16}. Here, $LT$, $RT$, $REL$ respectively refer to left term, right term and relationship between two terms. While tuples are efficient for storage, they are incompetent to represent relationships with more than $2$ terms. \citeauthor{DBLP:conf/semeval/WangLJWZZ20} create a sentence-level commonsense dataset, which validates the integrity of commonsense in real context. 

\subsection{Commonsense Reasoning with PLMs}

Large-scale pre-trained language models like BERT have drawn the most attention from the NLP community since their introduction. PLMs show their potential to significantly boost performance on NLP tasks across fields. Since PLMs have been trained on a large-scale corpus to learn interdependency between components, mining from PLMs for commonsense knowledge becomes a new method to create knowledge databases \citep{DBLP:conf/emnlp/PetroniRRLBWM19,DBLP:conf/acl/AlghanmiAS21,DBLP:conf/eacl/KassnerDS21}. LAMA \cite{DBLP:conf/emnlp/PetroniRRLBWM19} makes the first try to gather knowledge from PLMs by generative prompts. Later works follow this process to provide partial information in the commonsense knowledge tuple and require PLMs to complete the rest of the tuple. 

The commonsense knowledge understanding of PLMs inspires researchers to directly apply PLMs for downstream inference without supervised fine-tuning. Commonsense question answering \citep{DBLP:conf/aaaiss/RoemmeleBG11,DBLP:conf/emnlp/ZellersBSC18,DBLP:conf/naacl/TalmorHLB19,DBLP:journals/corr/abs-2201-05320,DBLP:journals/corr/abs-2004-13831} is commonly used to test the zero-shot inference ability of PLMs. Similar to commonsense reasoning, prompts are applied to transform the question-answer pair to a syntactically plausible sentence. PLM-based perplexity is calculated for those transformed sentences and the sentence with the lowest perplexity is used to select the corresponding question-answer pair \cite{DBLP:journals/corr/abs-1806-02847,DBLP:conf/aaai/BosselutBC21,DBLP:conf/acl/TamborrinoPPVN20}. Besides a direct reasoning on answer candidates, researchers have also tried to sample extra candidates from generators and use pre-trained semantic similarity evaluator for answer selection. \citep{DBLP:conf/emnlp/ShwartzWBBC20,DBLP:conf/acl/NiuHLC0H20,DBLP:conf/aaai/BosselutBC21}

Current mainstream PLMs, BERT or GPT2, apply the conventional perplexity metric to use the probability of generating components based on the context. This will incorporate lexical properties like word frequency as perturbance to the inference. Based on the nature of commonsense reasoning, we propose a pre-trained discriminator, like ELECTRA, to be an alternative for better performance. 

\section{PLM-based Metric}

\subsection{Casual Language Model}

GPT2 is a PLM for text generation, which can also be applied for inference based on perplexity of selection candidates. The training objective, CLM, is optimized based on context-based next-word prediction. 

\begin{equation*}
\centering
\begin{aligned}
\mathcal{L} \triangleq \textrm{CELoss}(\textrm{PLM}_\theta(w_{1:i-1}), \textrm{One-hot}(w_{i}))
\end{aligned}
\end{equation*} 

\noindent where CELoss is the cross-entropy loss, and One-hot refers to the one-hot encoding. $\theta, w$ respectively refer to PLM parameters and words. The inference procedure also takes next-word prediction for perplexity ($\textit{PPL}$) calculation.

\begin{equation*}
\centering
\begin{aligned}
p_i = p(w_i|\textrm{PLM}_\theta, w_{1:i-1})\\ \textit{PPL} = \frac{1}{n}\sum_{i=1}^n(-\log(p_i))
\end{aligned}
\end{equation*}  

\noindent where $n$ is the length of the sentence. GPT2 calculates $\textit{PPL}$ by scoring answer choices and selecting a candidate with the lowest perplexity. 

\subsection{Masked Language Model}

MLM is the training objective for most bidirectional PLMs like BERT and RoBERTa \cite{DBLP:journals/corr/abs-1907-11692}. MLM is similar to CLM as it also uses word retrieval as the training objective. The difference is MLM leverages the bidirectional context for the prediction. 

\begin{equation*}
\centering
\begin{aligned}
\mathcal{L} \triangleq \textrm{CELoss}(\textrm{PLM}_\theta(w_{1:i-1;i+1:n}), \textrm{One-hot}(w_{i}))
\end{aligned}
\end{equation*} 

Likewise, the inference step for MLM is revised as follows:

\begin{equation*}
\centering
\begin{aligned}
p_i = p(w_i|\textrm{PLM}_\theta, w_{1:i-1;i+1:n})
\end{aligned}
\end{equation*}  

\subsection{Replaced Token Detection}

RTD differs from the word retrieval-targeted training procedure above as it sets binary classification as the objective. The PLM involves a discriminator which discerns replaced words in the sentence following an adversarial architecture.

\begin{equation*}
\centering
\begin{aligned}
\mathcal{L} \triangleq \textrm{BCELoss}(\textrm{PLM}_\theta(w_{1:n}), f_B(w_i))
\end{aligned}
\end{equation*} 

\noindent where $f_B$ is a Boolean function that returns whether $w_i$ is corrupted by the replacement or not. 

We then introduce the Non-Replacement Confidence ($\textit{NRC}$) metric for confidence evaluation. 
 
\begin{equation*}
\centering
\begin{aligned}
& p_i = \textrm{PLM}_\theta(w_{1:n})\\
& \textit{NRC} = \frac{1}{n}\sum_{i=1}^n(-\log(p_i))
\end{aligned}
\end{equation*} 

\subsection{Metric Comparison}

$\textit{PPL}$ and $\textit{NRC}$ are both calculated based on negative log probability. While $\textit{PPL}$ evaluates the existing probability of a sentence, $\textit{NRC}$ reflects the confidence of contextual integrity. Thus, lower $\textit{PPL}$ and higher $\textit{NRC}$ on legal language indicate more human-like choice. 

Commonsense reasoning expects to understand the underlying interdependency between abstract concepts rather than their lexical properties. Thus, evaluating the confidence in the piece of commonsense knowledge should include not only words in the original sentence but their contextual synonyms as well.

\begin{equation*}
\centering
\begin{aligned}
p_{CS}(w_{1:n}) = \sum_{w\in \textrm{syn}(w_i)}p(C_i)p(w|C_i)
\end{aligned}
\end{equation*}  

\noindent where $p_{CS}$ is the \textbf{c}ommon\textbf{s}ense-targeted confidence. $C_i = w_{1:i-1;i+1:n}$ refers to the context for $w_i$ and syn returns the contextual synonyms of $w_i$. As $w_i \in \textrm{syn}(w_i)$, $p_{CS}(w_{1:n}) > p(w_{1:n}) = PPL$ when the number of synonym candidates is more than $1$, indicating that perplexity always underestimates the commonsense-targeted confidence. The underestimation becomes more severe when $w_i$ is a low-frequency word. Furthermore, as $\sum_{w\in dict}(p(w)) = 1$ ($dict$ is the whole dictionary for candidate selection), the correlation between confidence on synonym candidates is $-1$, which is contradicted to the fact that synonym supports each other for validation. 

In contrast, $\textit{NRC}$ does not require all candidates to share the distribution but evaluates individual confidence in each candidate. Thus, there is no underlying synonym candidate that leads to an underestimation or bias toward high-frequency words. The individual evaluation also changes the correlation between synonym candidates to positive as PLMs project contextually similar components to near positions in the latent space \cite{DBLP:conf/naacl/DevlinCLT19}. Theoretically, $\textit{NRC}$ is a more competent metric for commonsense reasoning than $\textit{PPL}$.

We also compare the time complexity of different metrics in Table~\ref{tab:tc}. Our NRC is as efficient as the CLM-based inference since token masking is not needed to calculate the metric, which limits the efficiency of MLM-based inference. 

\begin{table}
\centering
\scalebox{1.0}{
\begin{tabular}{lc}
\toprule
Metric & Time Complexity\\
\midrule
PPL$_{\textrm{CLM}}$ & $O(1)$ \\
PPL$_{\textrm{MLM}}$ & $O(n)$ \\
NRC & $O(1)$ \\
\bottomrule
\end{tabular}
}
\caption{Time complexity of different PLM-based metrics. The complexity counts the number of PLM forwarding.} 
\label{tab:tc}
\end{table}

\section{Commonsense Reasoning}

We test the PLMs on resources for commonsense reasoning before applying them to question answering. To mitigate the unfair comparison caused by the scale of parameters, this paper compares among large models with the same number of layers and hidden sizes, namely \textbf{BERT}$_\textrm{Large}$, \textbf{RoBERTa}$_\textrm{Large}$, \textbf{GPT2}$_\textrm{Medium}$ and \textbf{ELECTRA}$_\textrm{Large}$\footnote{\href{https://huggingface.co/google/electra-large-discriminator}{https://huggingface.co/google/electra-large-discriminator}} ($24$-layer, $1024$-hidden size). We also include \textbf{GPT2}$_\textrm{XLarge}$ ($48$-layer, $1600$-hidden size) for further comparison. Towards a strict unsupervised inference, we do not use any development dataset for hyperparameter selection.

\begin{table}
\centering
\scalebox{0.90}{
\begin{tabular}{lccc}
\toprule
Metric & ConceptNet & SemEval$_{\textrm{A}}$ & SemEval$_{\textrm{B}}$\\
\midrule
PPL$_{\textrm{GPT2-XL}}$ & 65.4 & 78.1 & 58.1 \\
\midrule
PPL$_{\textrm{GPT2-M}}$ & 49.6 & 50.1 & 40.3 \\
PPL$_{\textrm{BERT}}$ & 66.2 & 76.2 & 54.4 \\
PPL$_{\textrm{RoBERTa}}$ & 69.9 & 79.9 & 62.4 \\
NRC & \underline{\textbf{71.2}} & \underline{\textbf{80.5}} & \underline{\textbf{64.3}} \\
\bottomrule
\end{tabular}
}
\caption{Experiment results on tuple and sentence-level commonsense reasoning. \textbf{Bold:} The best performance on the dataset. \underline{Underline:} The result is significantly better than the second-best result. ($\alpha=0.01$)} 
\label{tab:cr}
\end{table}

\subsection{Tuple-level Probing}

\paragraph{ConceptNet\footnote{\href{https://home.ttic.edu/~kgimpel/commonsense.html}{https://home.ttic.edu/~kgimpel/commonsense.html}}} uses deep neural networks to retrieve commonsense candidates from corpus, which are validated by human annotators. Its training dataset contains more than $600,000$ tuples with different confidences. Its test dataset requires models to discern between true commonsense tuples and adversarial fake ones.

We follow LAMA \cite{DBLP:conf/emnlp/PetroniRRLBWM19} to create prompts\footnote{All prompts in our experiments can be found in Appendix~\ref{apdx:prompt}} for tuples in the test dataset that can be directly represented by natural languages. Then, we differentiate the prompts by PLM-based metrics and use accuracy to evaluate the results. 

Our experiment results are presented in Table~\ref{tab:cr}, NRC significantly outperforms both CLM and MLM-based PPL on commonsense tuple reasoning. Considering that transformed tuple relationships are simple and unified in syntactic structures, the discriminating ability is attributed to the understanding of commonsense. Thus, the results are convincing evidence for the superiority of NRC in commonsense validation. 

\subsection{Sentence-level Probing}

\paragraph{SemEval2020\footnote{\href{https://github.com/wangcunxiang/SemEval2020-Task4-Commonsense-Validation-and-Explanation}{https://github.com/wangcunxiang/SemEval2020-Task4-Commonsense-Validation-and-Explanation}}} collects natural language statements related to commonsense expression. We experiment on two reasoning subtasks. \textbf{A:} Select a statement that is against the commonsense. \textbf{B:} Select a reason for why the statement is against the commonsense. We continue evaluating and selecting statements and explanations according to different metrics.  

As the results in Table~\ref{tab:cr}, NRC is verified to perform significantly better than PPL on both differentiating and explanation, validating the superior evaluating capability of sentence-level commonsense of NRC. PPL$_\textrm{RoBERTa}$ is a competitive metric for differentiating since most statements use basic vocabulary in high frequency. Also, negative cases in SemEval are very anti-commonsense, which restrains the underestimation effect of PPL. When it comes to explanation, the gap between NRC and PPL$_\textrm{RoBERTa}$ becomes more significant since explanation requires a more complex inference ability. The comparison of sentence-level commonsense reasoning supports NRC to be a more competent metric for commonsense reasoning (differentiating and explanation) than PPL. 

\section{Commonsense Question Answering}

For commonsense reasoning, we are interested in not only how well models understand commonsense but also how well models leverage the understanding for downstream inference. Commonsense question answering is a commonly used downstream task for the practice of commonsense understanding. We also include sampling-based baselines\footnote{These methods generate many answer candidates from GPT2 to support the selection. They are more complex and time-consuming.} (Self-Talk \cite{DBLP:conf/emnlp/ShwartzWBBC20}, CGA \cite{DBLP:conf/aaai/BosselutBC21}, SEQA \cite{DBLP:conf/acl/NiuHLC0H20}) and other strong baselines to see if NRC achieves state-of-the-art performance.

\begin{table}
\centering
\scalebox{1.0}{
\begin{tabular}{llccc}
\toprule
Method & Trg & CSQA & ARC$_\textrm{E}$ & ARC$_\textrm{C}$\\
\midrule
Self-Talk & - & 32.4 & - & - \\
\midrule
\multirow{2}*{PPL$_{\textrm{GPT2-XL}}$} & A & 40.0 & 48.9 & 28.7 \\
& QA & 42.2 & 51.0 & 28.8 \\
\midrule
\midrule
\multirow{2}*{PPL$_{\textrm{GPT2-M}}$} & A & 34.9 & 42.5 & 26.5 \\
& QA & 35.7 & 43.9 & 26.9 \\
\midrule
\multirow{3}*{PPL$_{\textrm{BERT}}$} & Q & 42.4 & 37.8 & 27.5 \\
& A & 30.7 & 34.8 & 25.3 \\
& QA & 35.0 & 37.2 & 24.7 \\
\midrule
\multirow{3}*{PPL$_{\textrm{RoBERTa}}$} & Q & 45.7 & 38.6 & 33.7  \\
& A & 31.2 & 33.8 & 27.7 \\
& QA & 40.0 & 37.7 & 31.9 \\
\midrule
\multirow{3}*{NRC}  & Q &49.5 & 47.4 & 36.8 \\
& A &47.4 & 47.3 & 37.1 \\
& QA &\underline{\textbf{51.8}} & \underline{\textbf{51.7}} & \underline{\textbf{38.4}} \\
\bottomrule
\end{tabular}
}
\caption{Experiment results on phrase selection.} 
\label{tab:ps}
\end{table}

\subsection{Phrase Selection}

\paragraph{CommonsenseQA\footnote{\href{https://www.tau-nlp.org/commonsenseqa}{https://www.tau-nlp.org/commonsenseqa}} (CSQA)} provides remarkable resources for commonsense-targeted question answering since it builds question-answer pairs based on ConceptNet. The annotators create adversarial choices based on the subgraphs in ConceptNet. Specifically, negative choices are sampled from terms related to the question in ConceptNet, making differentiating confusing for models without strong commonsense understanding. 

\paragraph{ARC\footnote{\href{https://allenai.org/data/arc}{https://allenai.org/data/arc}}} is a commonsense question answering challenge that also selects phrases for science questions. The difficulty of questions is in grade-school level and the dataset is split into the easy part (ARC$_\textrm{E}$) and the challenging part (ARC$_\textrm{C}$). 

We follow previous works \citep{DBLP:conf/emnlp/ShwartzWBBC20,DBLP:conf/acl/NiuHLC0H20} to calculate the metrics on different targeted components (Question (Q), Answer (A), Question+Answer (QA)) for inference. The selection results in depicted in Table~\ref{tab:ps}. NRC outperforms PPL based on PLM on the same scale by a large margin ($6.1$, $7.8$, $4.7$ accuracy score), indicating NRC to be also superior in using commonsense for inference. For the easy part of ARC (ARC$_\textrm{E}$), large-scale models like GPT2$_\textrm{XL}$ seem to be able to compensate for bias in metric. However, when the questions become more challenging in ARC$_\textrm{C}$, the gap again reaches about $10.0$ accuracy scores, showing the inherent differences between NRC and PPL in commonsense reasoning ability. 

\subsection{Sentence Selection}

\begin{table}
\centering
\scalebox{1.0}{
\begin{tabular}{llcc}
\toprule
Method & Trg & COPA & Swag\\
\midrule
Self-Talk & - & 68.6 & - \\
CGA & - & 72.2 & - \\
SEQA & - & 79.4 & - \\
\midrule
\multirow{2}*{PPL$_{\textrm{GPT2-XL}}$} & A & 73.6 & 65.3 \\
& QA & 71.6 & 64.9 \\
\midrule
\midrule
\multirow{2}*{PPL$_{\textrm{GPT2-M}}$} & A & 68.4 & 59.7 \\
& QA & 66.6 & 59.1 \\
\midrule
\multirow{3}*{PPL$_{\textrm{BERT}}$} & Q & 64.2 & 44.5 \\
& A & 61.2 & 63.4 \\
& QA & 64.2 & 64.1 \\
\midrule
\multirow{3}*{PPL$_{\textrm{RoBERTa}}$} & Q & 70.6 & 48.1 \\
& A & 68.4 & 71.0 \\
& QA & 75.2 & 74.5 \\
\midrule
\multirow{3}*{NRC}  & Q & \underline{\textbf{82.6}} & 24.5 \\
& A & 71.2 & \underline{\textbf{77.4}} \\
& QA & 78.4 & 75.4 \\
\bottomrule
\end{tabular}
}
\caption{Experiment results on sentence selection.} 
\label{tab:ss}
\end{table}

\paragraph{COPA\footnote{\href{https://people.ict.usc.edu/~gordon/copa.html}{https://people.ict.usc.edu/~gordon/copa.html}}} is a simple commonsense-targeted question answering dataset. COPA is interested in choosing the cause or effect of a sentence. 

\paragraph{Swag\footnote{\href{https://rowanzellers.com/swag/}{https://rowanzellers.com/swag/}}} is a large-scale commonsense question answering dataset with more than $20,000$ test data. The question is formulated as entailment that aims to satisfy the contextual integrity in commonsense. 

Experiment results on sentence selection are presented in Table~\ref{tab:ss}. NRC again shows superior performance over PPL ($7.4$ on COPA, $2.9$ on Swag), validated by the large Swag dataset. This verifies the superiority of NRC in the application of phrase and sentence-level commonsense understanding for downstream inference. Compared to sampling-based methods, the outstanding performance of NRC also boosts state-of-the-art. The question part of Swag is not very useful for NRC probably because these questions are not dependent on the answer choices on the view of ELECTRA, which prefers to use the answer part of this dataset for inference. But when evaluating the whole question-answer pair (QA), NRC always performs better than PPL. 

\subsection{Context-based Selection}

\begin{table}
\centering
\scalebox{1.0}{
\begin{tabular}{llccc}
\toprule
Method & Trg & SCT & SQA & CQA\\
\midrule
Self-Talk & - & 70.4 & 47.5 & 36.1 \\
CGA & - & 71.5 & 45.4 & 42.2 \\
SEQA & - & 83.2 & 47.5 & 56.1 \\
\midrule
\multirow{2}*{PPL$_{\textrm{GPT2-XL}}$} & A & 70.6 & 41.4 & 35.5 \\
& QA & 71.5 & 41.4 & 31.1 \\
\midrule
\midrule
\multirow{2}*{PPL$_{\textrm{GPT2-M}}$} & A & 54.0 & 35.6 & 27.0 \\
& QA & 55.4 & 35.4 & 18.2 \\
\midrule
\multirow{3}*{PPL$_{\textrm{BERT}}$} & Q & 63.5 & 35.7 & 32.9 \\
& A & 58.2 & 35.4 & 30.7 \\
& QA & 61.2 & 38.5 & 29.6 \\
\midrule
\multirow{3}*{PPL$_{\textrm{RoBERTa}}$} & Q & 61.5 & 37.1 & 38.6 \\
& A & 67.3 & 41.4 & 36.1 \\
& QA & 71.7 & 41.5 & 36.5 \\
\midrule
\multirow{3}*{NRC}  & Q & 65.0 & 42.8 & 41.2 \\
& A & 74.7 & 43.0 & 41.9 \\
& QA & \underline{\textbf{77.1}} & \underline{\textbf{45.1}} & \underline{\textbf{44.3}} \\

\bottomrule
\end{tabular}
}
\caption{Experiment results on context-based selection.} 
\label{tab:cbs}
\end{table}

\paragraph{StoryClozeTest\footnote{\href{https://cs.rochester.edu/nlp/rocstories/}{https://cs.rochester.edu/nlp/rocstories/}} (SCT)} is a story entailment dataset that collects $5$-sentence stories with multiple ending candidates. We use the first three sentences as context and the fourth as question. 

\paragraph{SocialiQA\footnote{\href{https://leaderboard.allenai.org/socialiqa/submissions/public}{https://leaderboard.allenai.org/socialiqa/submissions/public}} (SQA)} contains questions about interactions of people in social activities. The context describes a social circumstance with related aspects, and the question asks the model to select a proper interaction. 

\begin{table*}
\centering
\scalebox{0.83}{
\begin{tabular}{lcccccccc}
\toprule
Method & CSQA & ARC$_\textrm{E}$ & ARC$_\textrm{C}$ & COPA & Swag & SCT & SQA & CQA\\
\midrule
PPL$_{\textrm{GPT2-M}}$ & 35.7 (0.0) & 42.8 (-1.1) & \underline{27.5} (0.6) & \underline{69.4} (1.0) & 59.3 (0.2) & 53.2 (-2.2) & 33.7 (-1.9) & 26.9 (-0.1) \\
PPL$_{\textrm{BERT}}$ & 42.1 (-0.3) & 36.3 (-1.5) & 27.1 (-0.4) & \underline{66.6} (2.2) & 63.5 (-0.6) & 63.0 (-0.5) & 36.7 (-1.8) & 32.1 (-0.8) \\
PPL$_{\textrm{RoBERTa}}$ & 45.0 (-0.7) & 37.3 (-1.8) & 33.2 (-0.5) & 74.4 (-0.8) & 73.2 (-1.3) & \underline{72.1} (0.4) & 41.2 (-0.3) & 38.6 (0.0) \\
NRC & \underline{52.3} (0.5) & 51.9 (0.2) & \underline{39.8} (1.4) & \underline{84.2} (1.6) & 74.6 (-2.8) & 76.6 (-0.5) & \underline{46.6} (1.5) & 44.5 (0.2) \\
\bottomrule
\end{tabular}
}
\caption{Effect of the removal of stop words. \underline{Underline:} The removal results in a significant improvement.} 
\label{tab:stop}
\end{table*}

\paragraph{CosmosQA\footnote{\href{https://wilburone.github.io/cosmos/}{https://wilburone.github.io/cosmos/}} (CQA)} is similar to COPA as it also asks the cause and effect of events. The difference is that CosmosQA provides an event background as the context for the question. Also, the answer of CosmosQA is longer than other datasets, which increases the difficulty for inference. 

As in Table~\ref{tab:cbs}, NRC outperforms PPL based on PLMs in the scale and the large-scale GPT2$_{\textrm{XLarge}}$ by a significant gap. On datasets with a long context (SCT and CQA), the gap become larger, reflecting the capability of NRC to understand the interdependency between terms in more complex contexts. On context-based selection, sampling-based method on GPT2$_{\textrm{XLarge}}$ still holds state-of-the-art, which indicates that larger-scale language models still encode more knowledge in the network with much more parameters. However, the generative nature limits the understanding of the knowledge and sampling is essential to generate multiple candidates to fully retrieve the knowledge from the network. We hold the belief that better performance and efficiency will be achieved by a larger-scale ELECTRA, which is left for future work. 

\section{Further Analysis}

\subsection{Source of Reasoning Ability}

\begin{table}
\centering
\scalebox{0.93}{
\begin{tabular}{lcccc}
\toprule
$\Delta W$ & PPL$_{\textrm{GPT2-M}}$ & PPL$_{\textrm{BERT}}$ & PPL$_{\textrm{RoBERTa}}$ & NRC\\
\midrule
0.00 & 35.7 & \bf 42.4 & \bf 45.7 & 51.8 \\
0.25 & 35.5 & 41.8 & 45.0 & 51.9 \\
0.50 & \bf 35.9 & 41.2 & 44.8 & \bf 52.2 \\
0.75 & 35.7 & 40.6 & 44.0 & 51.7 \\
1.00 & 35.7 & 40.2 & 43.6 & 51.7 \\
\bottomrule
\end{tabular}
}
\caption{Benefits of extra weights on question concept. \textbf{Bord:} Best performance of each PLM.} 
\label{tab:qcw}
\end{table}

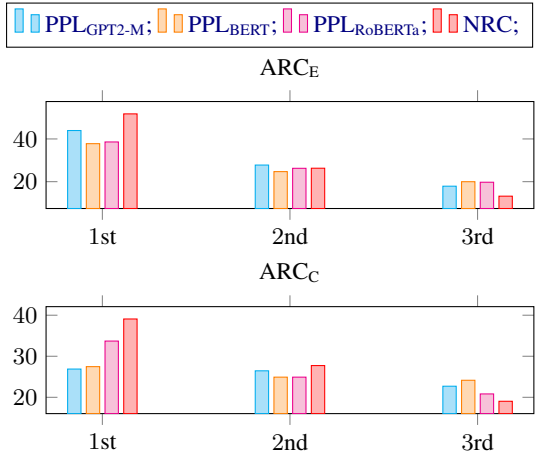
\begin{figure}
\begin{center}
    \ref{namedign}
    \centering
    \begin{tikzpicture}
    \small
    \begin{axis}[
        legend columns=-1,
        legend entries={PPL$_{\textrm{GPT2-M}}$;,PPL$_{\textrm{BERT}}$;,PPL$_{\textrm{RoBERTa}}$;,NRC;},
        legend to name=namedign,
        legend style={font=\small},
        ybar,
        enlargelimits=0.15,
        width=8cm,
        height=3.0cm,
        bar width=5pt,
        xticklabels={$1$st, $2$nd, $3$rd},
        xtick=data,
        style={font=\small},
        title=ARC$_{\textrm{E}}$
    ]
    \addplot[draw=cyan, fill=cyan!30] coordinates{
        (1, 43.94)
        (2, 27.78)
        (3, 17.89)
    };
    \addplot[draw=orange, fill=orange!30] coordinates{
        (1, 37.79)
        (2, 24.71)
        (3, 19.99)
    };
    \addplot[draw=magenta, fill=magenta!30] coordinates{
        (1, 38.59)
        (2, 26.26)
        (3, 19.74)
    };
    \addplot[draw=red, fill=red!30] coordinates{
        (1, 51.73)
        (2, 26.3)
        (3, 13.22)
    };
    \end{axis}
    \end{tikzpicture}
    \begin{tikzpicture}
    \small
    \begin{axis}[
        ybar,
        enlargelimits=0.15,
        width=8cm,
        height=3.0cm,
        bar width=5pt,
        xticklabels={$1$st, $2$nd, $3$rd},
        xtick=data,
        style={font=\small},
        title=ARC$_{\textrm{C}}$
    ]
    \addplot[draw=cyan, fill=cyan!30] coordinates{
        (1, 26.88)
        (2, 26.45)
        (3, 22.7)
    };
    \addplot[draw=orange, fill=orange!30] coordinates{
        (1, 27.47)
        (2, 24.91)
        (3, 24.15)
    };
    \addplot[draw=magenta, fill=magenta!30] coordinates{
        (1, 33.7)
        (2, 24.91)
        (3, 20.82)
    };
    \addplot[draw=red, fill=red!30] coordinates{
        (1, 39.08)
        (2, 27.73)
        (3, 19.03)
    };
    \end{axis}
    \end{tikzpicture}
    \caption{Ranks of PLM-based selection on easy and challenging ARC.}
    \label{fig:arc}
\end{center}
\end{figure}

\paragraph{Stop Word} For models that leverage commonsense to infer, stop words actually add noise to the inference as humans rarely use them for commonsense reasoning. Thus, we remove the scores calculated on stop words and test whether this will boost the performance of PLM-based metrics. We sample stop words from the pool provided by SpaCy to set articles and pronouns as stop words. 

The effect of the removal of stop words is shown in Table~\ref{tab:stop}. NRC benefits the most from the removal of stop words, which leads to (significant) improvement on $6$ ($4$) out of $8$ datasets. We thus conclude that NRC better takes advantage of the non-trivial components to infer.

\paragraph{Question Concept} CommonsenseQA annotates the commonsense-related phrase in each question. These phrases are connected to answer candidates in the ConceptNet. For models adept at using commonsense for inference, a higher weight on the phrase should be beneficial for the inference. We thus add extra weights ($\Delta W$) and investigate the effect on different metrics. 

Table~\ref{tab:qcw} presents the effect of concentration on question concepts. Extra weight negatively contributes to the inference of MLM-based PLMs, indicating that they are unsuccessful in applying commonsense understanding to infer. As the negative candidates are also sampled from the neighbors of the question concept in the ConceptNet, these models are confused by the ambiguity. Compared to PPL$_{\textrm{GPT2-M}}$, ELECTRA-based NRC benefits more from the extra weight. This verifies our claim that a discriminator better models commonsense knowledge and leverages them to infer.

\subsection{Specific Statistics}

\begin{figure}
    \centering
    \includegraphics[width=0.45\textwidth]{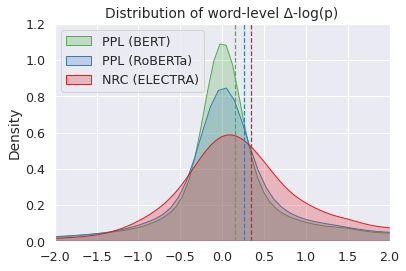}
    \caption{Distribution of the word-level differences in log probability. \textbf{Dashed line:} Average difference.}
    \label{fig:distribution}
\end{figure}

\begin{figure}
\begin{center}
    \ref{named}
    \centering
    \begin{tikzpicture}
    \small
    \begin{axis}[
        legend columns=-1,
        legend entries={PPL$_{\textrm{BERT}}$;,PPL$_{\textrm{RoBERTa}}$;,NRC;},
        legend to name=named,
        legend style={font=\small},
        width=8.0cm,
        height=4.0cm,
        xmin=1, xmax=45,
        minor tick num=1,
        grid=both,
        grid style=dashed,
        ]
        \addplot[color=teal,mark=*,mark size=1.5pt,] coordinates{
        (1, 0.23)
        (2, 0.2)
        (3, 0.22)
        (4, 0.14)
        (5, -0.053)
        (6, 0.18)
        (7, 0.18)
        (8, 0.19)
        (9, -0.23)
        (15, -0.0033)
        (25, 0.043)
        (35, 0.036)
        (45, -0.0051)
        };
        \addplot[color=cyan,mark=*,mark size=1.5pt,] coordinates{
        (1, 0.42)
        (2, 0.24)
        (3, 0.32)
        (4, 0.24)
        (5, 0.12)
        (6, 0.28)
        (7, 0.027)
        (8, 0.063)
        (9, 0.055)
        (15, 0.072)
        (25, 0.13)
        (35, 0.097)
        (45, 0.0017)
        };
        \addplot[color=red,mark=*,mark size=1.5pt,] coordinates{
        (1, 0.52)
        (2, 0.44)
        (3, 0.35)
        (4, 0.36)
        (5, 0.37)
        (6, 0.38)
        (7, 0.027)
        (8, 0.32)
        (9, 0.47)
        (15, 0.35)
        (25, 0.27)
        (35, 0.19)
        (45, 0.12)
        };
    \end{axis}
    \end{tikzpicture}
    \label{fig:sem_np}
    \caption{Relationship between word frequency and its contribution to the inference.}
    \label{fig:freq}
\end{center}
\end{figure}
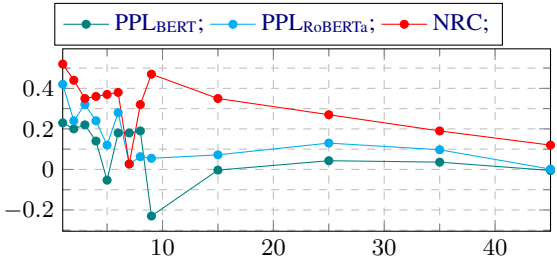

\paragraph{Rank of the Choice} The accuracy only counts the matching between the golden answer and the first-rank choice. We show the ranking distribution of selected answers in Table~\ref{fig:arc} to further investigate the inference results. On the easy subsets of ARC, there does not exist a prominent advantage of NRC according to the second-rank choice rates. But when the questions become challenging, the rate of golden answers in the second rank rises, reflecting the superior capability of NRC in more challenging question answering. 

\paragraph{Difference Distribution} We depict the difference distribution of log probability on COPA in Figure~\ref{fig:distribution}. We compare the predicted probability on the question part when it is attached by a positive or negative choice. Words are viewed as voters whose contribution to the positive choice is reflected by the difference. PPL$_{\textrm{GPT2-M}}$ is not included since the answer makes no difference for the question component for unidirectional PLMs. Compared to NRC, PPL difference is more likely to distribute around $0.0$, indicating its lower differentiating ability. Also, the average value of NRC difference is greater than PPL difference, again supporting the stronger inference ability of NRC. 

\paragraph{Contribution v.s. Frequency} We continue studying on the contributions of word voters. We count the frequency of words in the COPA dataset and show the relationship with their contributions in Table~\ref{fig:freq}. On words with frequency $<10$, NRC evaluation provides more positive and stable support to the right answer. The results verify our claim that NRC better evaluates the semantics of low-frequency words. The advantage of NRC over PPL decreases when the frequency rises, but NRC still holds the superiority as high-frequency words also suffer from the confidence taken by synonyms. 

\subsection{Case Study}

\begin{figure}
    \centering
    \includegraphics[width=0.43\textwidth]{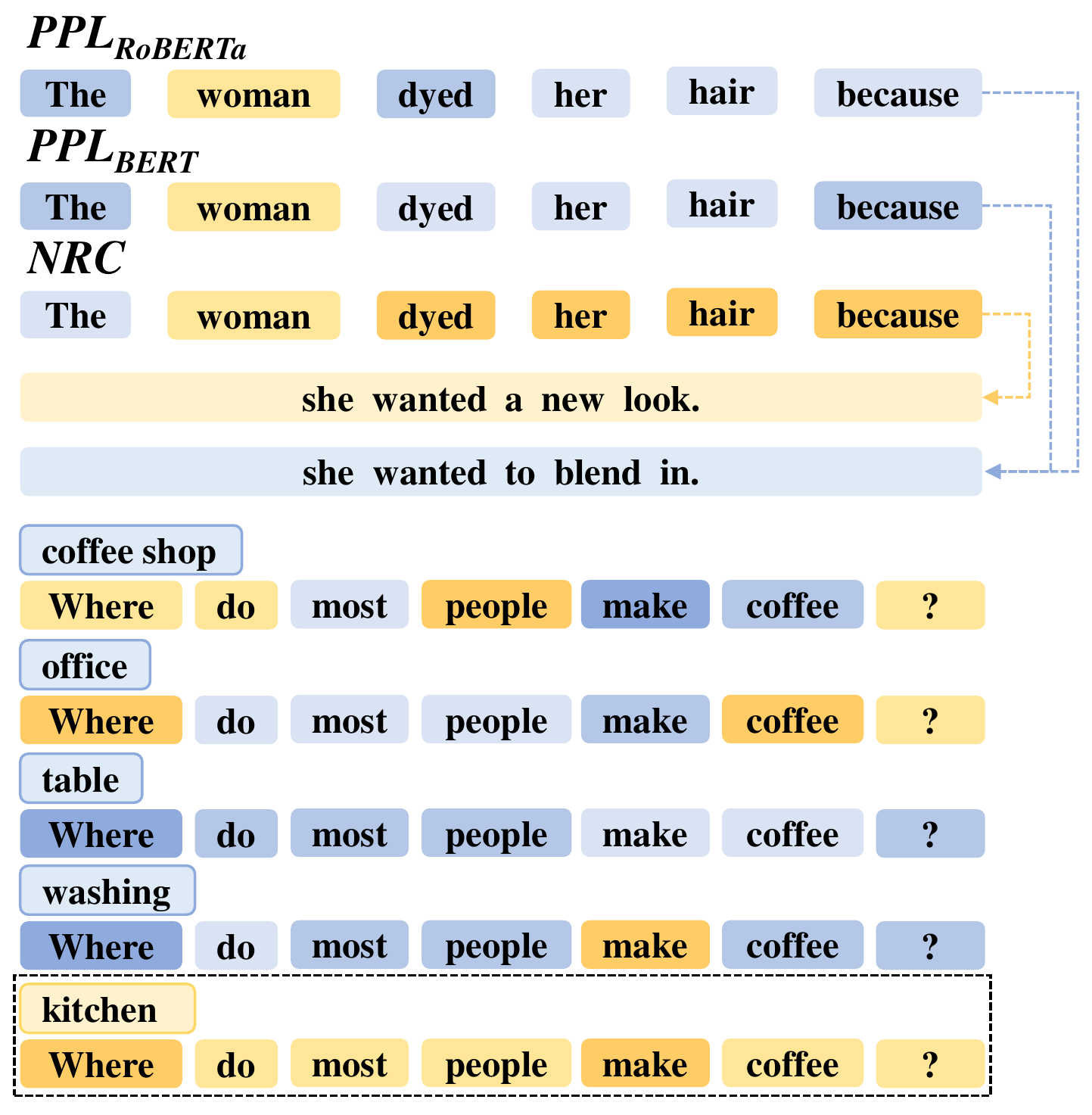}
    \caption{A case study on the inference of NRC.}
    \label{fig:case}
\end{figure}

We provide cases to specify the observation in statistics. The first case on COPA shows the limited understanding of PPL on the low-frequency phrase \textit{dyed her hair}. NRC instead successfully leverages the semantics of the phrase to select the right answer. The second case on CommonsenseQA shows NRC to infer based on \textit{Where} and \textit{make coffee} and selects the answer supported by both key phrases, verifying its inference to be highly explainable. 

\section{Conclusion}

This paper suggests replacing perplexity with confidence to make the commonsense-targeted reasoning. We investigate the bias in the application of perplexity for inference. We propose a superior alternative, RTD-based non-replacement confidence, for better evaluation. Experiments on a wide range of commonsense reasoning and question answering datasets provide comprehensive evidence and analysis for the superiority of NRC. 

\section{Limitation}

The nature of discriminator enables NRC to infer without the perturbance caused by synonyms and their word frequency, which supports the strong performance of NRC in commonsense reasoning. However, when we want a specific term during commonsense knowledge mining, the discriminator is not able to return it. Thus, we recommend combining the discriminator and a generator like GPT2 for commonsense knowledge mining. For instance, after we sample several terms like \textit{warm}, \textit{cold}, \textit{chilly} from the prompt \textit{The weather is \_}, we can use the pre-trained discriminator to re-rank the confidence. We thus leave the application of pre-trained discriminators to mine commonsense knowledge for future works. 

\bibliography{anthology,custom}
\bibliographystyle{acl_natbib}

\clearpage

\appendix

\section{Dataset Statistics}

The statistics of datasets in our experiments is presented in Table~\ref{tab:stat}.

\begin{table}
\centering
\scalebox{1.0}{
\begin{tabular}{lccccc}
\toprule
Dataset & $N_{\textrm{Inst}}$ & $N_{\textrm{A}}$ & $L_{Q}$ & $L_{A}$ & $L_{C}$ \\
\midrule
CSQA & $1140$ & $5$ & $13.2$ & $1.5$ & - \\
ARC$_\textrm{E}$ & $2376$ & $4$ & $19.6$ & $3.7$ & - \\
ARC$_\textrm{C}$ & $1172$ & $4$ & $20.6$ & $5.0$ & - \\
COPA & $500$ & $2$ & $6.1$ & $5.0$ & - \\
Swag & $20005$ & $4$ & $12.4$ & $11.2$ & - \\
SCT & $1571$ & $2$ & $8.9$ & $7.4$ & $26.4$  \\
SQA & $3525$ & $3$ & $11.2$ & $5.0$ & $19.6$ \\
CQA & $6510$ & $4$ & $12.0$ & $7.4$ & $43.9$ \\
\bottomrule
\end{tabular}
}
\caption{Statistics of datasets in our experiments. $N_{\textrm{inst}},N_A$: Number of instances and answer candidates. $L_Q,L_A,L_C$: Average length of question, answer and context.} 
\label{tab:stat}
\end{table}

\section{Prompts}
\label{apdx:prompt}

The prompts we used in experiments on ConceptNet are listed in Table~\ref{tab:prompt}. For SemEval$_{\textrm{B}}$, we use the prompt \textit{"A" is not true because B.} to select an explanation for unreal commonsense expression. Prompts for question answering follow the previous configuration \cite{DBLP:conf/acl/NiuHLC0H20} by attach the answer after the question.

\begin{table}
\centering
\scalebox{1.0}{ 
\begin{tabular}{lp{3cm}}
\toprule
Rel. & Prompt\\
\midrule
IsA & A is a B . \\
CapableOf & A is able to B .\\
NotCapableOf & A is unable to B .\\
UsedFor & A is used to B .\\
MadeOf & A is made of B .\\
PartOf & A is part of B .\\
HasAttribute & A is very B .\\
HasA & A has a B .\\
\bottomrule
\end{tabular}
}
\caption{Prompts used in experiments on ConceptNet. } 
\label{tab:prompt}
\end{table}

\end{document}